# Hallucination Stations
## On Some Basic Limitations of Transformer-Based Language Models

Varin Sikka  
Stanford University

Vishal Sikka  
VianAI Systems

**Abstract**

In this paper we explore hallucinations and related capability limitations in LLMs and LLM-based agents from the perspective of computational complexity. We show that beyond a certain complexity, LLMs are incapable of carrying out computational and agentic tasks or verifying their accuracy.

**Introduction**

With widespread adoption of transformer-based language models ("LLMs") in AI, there is significant interest in the limits of LLMs' capabilities, specifically so-called "hallucinations", occurrences in which LLMs provide spurious, factually incorrect or nonsensical [1, 2] information when prompted on certain subjects. Furthermore, there is growing interest in "agentic" uses of LLMs - that is, using LLMs to create "agents" that act autonomously or semi-autonomously to carry out various tasks, including tasks with applications in the real world. This makes it important to understand the types of tasks LLMs can and cannot perform. We explore this topic from the perspective of the computational complexity of LLM inference. We show that LLMs are incapable of carrying out computational and agentic tasks beyond a certain complexity, and further that LLMs are incapable of verifying the accuracy of tasks beyond a certain complexity. We present examples of both, then discuss some consequences of this work.

**Computational Complexity of LLMs, and its Implications**

As described in numerous introductory documents, LLMs have a vocabulary of a size $n$ set of tokens $t_1, t_2, ..., t_n$ (A token is a separable component of a word, such as "walk" and "ed" in "walked".) Their basic operation is to take a string of tokens from this vocabulary, and produce a string of tokens as output. The output string is produced one token at a time, and the input string and the generated tokens are both used to produce the next token.

An LLM performs several steps in its computation. While these steps can have some differences depending on the architecture of the LLM, it is well-established that the conventional self-attention mechanism processes an input consisting of $N$ tokens, each represented as a $d$-dimensional vector, with a computational time complexity of $O(N^2 \cdot d)$ to produce the resulting sequence [3, 4]. The other operations in it have complexity that is linear (i.e. $O(N)$) or less. To be more precise, the computing self-attention is $O(N^2.d + N.d^2)$ because some operations in it are quadratic in $d$ and the others are quadratic in $N$. There can of course be cases where $N$ is smaller than $d$, however, in general $N$ tends to be much larger, and modern LLMs have increasing context window sizes, e.g. Gemini [5] and GPT4.1 [6] have context windows of 1 million tokens or higher. In this paper therefore we assume that the LLM's overall computational complexity is $O(N^2 \cdot d)$. This means that for an input string of length $N$, the LLM performs approximately $N^2 \cdot d$ floating-point operations, regardless of the specific input. For example, the `Llama-3.2-3B-Instruct` model on our system setup, when given any input string of 17 tokens (such as "`You are a helpful assistant, please explain the following concept in detail: renewable energy`" or "`Can you find a number that when squared it equals four times its double minus twelve`") always runs a total of 109243372873 floating point operations. In [7] we walk through the computations performed in more detail and share our code.

Our intuition in this paper is: if there is an input string that expresses a task with computational complexity is higher than $O(N^2 \cdot d)$, then an LLM cannot correctly carry out that task. We first elaborate on this intuition with a few examples.

**Example 1:** Token Composition

Consider the task: "Given a size n set of tokens $\{t_1, t_2, \ldots, t_n\}$, list every string of length $k$ tokens." Computing this task takes $O(n^k)$ time. For $n = 2$ it takes $O(2^k)$ time. So this task, or any variant of it (e.g. "How many sentences of length $k$ are possible by combining the tokens in the set tokens $\{t_1, t_2, \ldots, t_n\}$?") cannot correctly be carried out by an LLM that executes in $O(N^2 \cdot d)$ time[1].

This is because a task requiring a certain minimum number of steps cannot, obviously, be correctly completed in a smaller number of steps. An LLM takes the input string "Given a set of tokens ... strings of length $k$ tokens" and proceeds to enumerate a string of tokens in succession, each token obtained as a result of a $O(N^2 \cdot d)$ calculation that picks the token with the highest likelihood of being next. However, this is not the same as carrying out the task at hand, which is to compute and enumerate the exponentially large number of possible sequences from the given set of tokens. Although $N$ is larger than $n$, for larger values of $n$ and $k$, $n^k$ will far exceed $O(N^2 \cdot d)$.

**Example 2:** Matrix Multiplication

The naive matrix multiplication algorithm involves the multiplication of two matrices by performing the dot product of the rows from the first matrix and the columns of the second matrix. Given two matrices $A$ and $B$, where $A$ is of size $m \cdot n$ and $B$ is of size $n \cdot p$, a matrix $C$ that is the product of $A$ and $B$ has size $m \cdot p$, where each element $C_{ij}$ is calculated as $C_{ij} = \left(\sum_{k=0}^{n-1}\right) A_{ik} \cdot B_{kj}$

We can write this algorithm as follows:

```
Initialize matrix C with dimensions m · p
For each row i in A:
        For each column j in B:
                Compute C_ij as  C_ij = (Σ_{k=0}^{n-1}) A_ik · B_kj
```

---

[1] $N$ here represents the number of tokens in the input string ("Given a size ... tokens."), whereas $n$ represents the size of the set that is within the prompt, which is always smaller than $N$ since the string expressing the $n$ token set is a substring of the input string, which includes additional tokens to express the task.

This algorithm takes $O(n^3)$ time, or more precisely, $O(m \cdot n \cdot p)$ time, as the first loop is run $m$ times, the second loop is run $p$ times, and within this inner loop the dot product operation has $n$ steps. For LLMs where $m$, $n$, and $p$ exceed its vocabulary size, the LLM will not be able to correctly carry out such a matrix multiplication algorithm.

Similarly, there are countless other tasks whose minimum computational complexity is $O(N^3)$ or higher. Fast-growing functions of cubic time or higher computational complexity frequently arise in real world scenarios. Examples include super-exponential functions like Ackermann Function (and its relevance to problems such as Petri Net reachability or Vector addition [8, 9], to Floyd-Warshall's algorithm for finding shortest paths between all pairs of nodes in a network (which has a computational complexity that is cubic in the number of nodes), to subset enumeration (which is generally exponential in the size of the set), several join operations in relational databases, computational fluid dynamics (e.g. Navier-Stokes [10]), and more.

**Example 3:** Agentic AI

For our third example, we consider agentic uses of LLMs. Agentic AI refers to AI systems capable of autonomous decision-making and goal-oriented behavior, where agents using LLMs carry out tasks (sometimes with none to minimal human intervention) [11, others]. Recently, interest in agentic AI has increased significantly due to its role in automating various tasks and most major technology companies have introduced products centered on agentic AI recently [12]. Such tasks can range from purely informational [13], to tasks that have side-effects in the real world, from making financial transactions [14] to purchasing products or services [15], from booking travel and managing hotel and restaurant reservations [16], and filing taxes or legal documents [17] to managing industrial equipment [18].

Given that agentic use of LLMs can go beyond informational activities to taking actions that affect lives and work in unprecedented ways, it is particularly important to examine their limitations. Our two earlier examples also apply to agentic use-cases, where agents are instructed to autonomously or semi-autonomously carry out tasks. Indeed each task given to an LLM can easily be cast as an agentic task. Tasks that AI agents are instructed to perform, can clearly have computational complexity beyond $O(N^2.d)$. Consequently LLM-based agents cannot correctly carry out those tasks either. In this vein, we will now explore if agents can be used to verify the correctness of another agent's solution to a given task. In the following, we will show that this is not possible either, because verification of a task is often harder than the task itself.

Let us consider the case of the Traveling Salesperson Problem (TSP), a computational task where solution verification can require exponential time. Given $n$ cities and a symmetric distance matrix, the problem is to verify whether a claimed route $R$ with total distance $D_R$ is the shortest among all possible routes. If the verification must be carried out without leveraging precomputed bounds, heuristics, or other approximations, then this requires $R$ to be compared against all $(n-1)!/2$ possible routes. This brute-force approach grows factorially with $n$, making the verification process significantly greater than our correctness threshold of $O(N^2.d)$. For instance, verifying a claimed solution for a 20-city TSP instance involves evaluating approximately $10^{17}$ candidate routes.

This phenomenon, beyond the TSP, applies to other real-world problems where exhaustive verification is required. Tasks such as vehicle routing in logistics, bin packing in warehouses, and scheduling tasks involving crew/staff or resources e.g. in airlines or services companies, or tasks that require solving the quadratic assignment problem [19] often also require exponential time for exact solutions. In formal verification of hardware and software systems, model checking often incurs state explosion, where the number of system states grows exponentially with the number of components or variables, making exhaustive verification computationally intractable for large designs [20]. We believe this case to be especially pertinent since one of the most prevalent applications of LLMs is to write and verify software [21].

These examples and their variants show that attempts by LLM-based agents to verify the correctness of tasks performed by other agents, will in general not work. Suppose $A_1$ and $A_2$ are two agents in the agentic AI sense — that is, agents that carry out tasks using an LLM. Let $A_1$ be tasked with executing a problem $P$ with a computational complexity of $O(n^3)$ or higher, where $n$ is included in the input prompt provided to $A_1$. Let $A_2$ be tasked with validating, i.e. verifying the correctness, of $A_1$'s solution for $P$. Since all of $A_2$'s operations are limited to $O(N^2 \cdot d)$ complexity (note once more the difference between $N$ and $n$), given that the inherent complexity of $P$, i.e. $O(n^3)$ or higher, exceeds $A_2$'s maximum computational complexity, it follows that, in general, $A_2$ cannot accurately verify the correctness of $A_1$'s solution to $P$. This is because any such verification procedure for $P$ would itself in general require at least $O(n^3)$ time complexity in order to function reliably.

With these examples in mind, we can now state the following theorem:

**Theorem 1:**

Given a prompt of length $N$, which includes a computational task within it of complexity $O(n^3)$ or higher, where $n < N$, an LLM, or an LLM-based agent, will unavoidably hallucinate in its response.

**Proof:**

Hartmanis and Stearns, in their seminal time-hierarchy theorem [22], showed that if $t_2(n)$ is an asymptotically larger function than $t_1(n)$ (e.g., $t_2(n) = n^2$ and $t_1(n) = n$), then there are decision problems solvable in $O(t_2(n))$ but not in $O(t_1(n))$. Consequently, any task that requires time greater than $O(N^2.d)$, such as the ones in our examples above, but indeed infinitely many such tasks, will not be correctly carried out by LLMs.

A corollary to this theorem is: there are tasks that can be given to LLM agents to perform, whose verification or whose check for accuracy or semantic properties, cannot be correctly performed by LLMs. To show this, it suffices to show that many verification tasks require computational complexity beyond $O(N^2.d)$. In more practical situations, as described earlier, countless tasks of polynomial and non-polynomial time complexity exist whose verification is worse than $O(N^2.d)$.

**Discussion**

Our argument, in essence, is: if the prompt to an LLM specifies a computation (or a computational task) whose complexity is higher than that of the LLM's core operation, then the LLM will in general respond incorrectly. There are prompts where the response from the LLM is necessarily wrong, and there are prompts where the response from the LLM may accidentally be correct, even though it carried out the task incorrectly. As mentioned earlier, we include both these cases within the broader category of hallucinations. Also, any LLM-based agent (i.e. in the Agentic AI sense) cannot correctly carry out tasks beyond the $O(N^2.d)$ complexity of LLMs. Further, LLMs or LLM-based agents cannot correctly verify the correctness of tasks beyond this complexity, and we have shared multiple examples of such real-world tasks. Although this was done to show practical applications of the theorem outlined above, it suffices to show that we can generate a prompt that simply instructs the LLM to perform any task it chooses involving X floating-point operations, where X is engineered to exceed the number of floating-point operations performed by that particular LLM in response, given the length of the prompt and the LLM's dimensionality and other properties. In a sense, one can think of an LLM's "intelligence" or intellectual ability, as bounded by this threshold.

This leads us to conclude that, despite their obvious power and applicability in various domains, extreme care must be used before applying LLMs to problems or use-cases that require accuracy, or solving problems of non-trivial complexity. Mitigating these limitations is an area of significant ongoing work, and various approaches are being developed, from composite systems [23] to augmenting or constraining LLMs with rigorous approaches [24, 25, 26]. In this light, it is important to also note that while our work is about the limitations of individual LLMs, multiple LLMs working together can obviously achieve higher abilities. Several bodies of work across the history of AI research, for example, both Simon's work on Sciences of the Artificial [27] and Minsky's work on The Society of Mind [28], argue that intelligence is a collective ability, that intelligence emerges from pieces that may not be intelligent.

One question that we have often heard recently is, do reasoning models overcome these limitations? While we will analyze this question more rigorously in subsequent work, intuitively we don't believe they do. Reasoning models, such as OpenAI's o3 and DeepSeek's R1, generate a large number of tokens in a "think" or "reason" step, before providing their response. So an interesting question is, to what extent does the generation of these additional tokens bridge the underlying complexity gap? Can the additional think tokens provide the necessary complexity to correctly solve a problem of higher complexity? We don't believe so, for two fundamental reasons: one that the base operation in these reasoning LLMs still carries the complexity discussed above, and the computation needed to correctly carry out that very step can be one of a higher complexity (ref our examples above), and secondly, the token budget for reasoning steps is far smaller than what would be necessary to carry out many complex tasks. Recent work by researchers at Apple [29] shows that reasoning models suffer from a "reasoning collapse" when faced with problems of higher complexity. They use "Towers of Hanoi" as their reference problem, a problem which (like some of our examples above) inherently requires an exponential time computation.

**Acknowledgements:**

The authors wish to thank: Prof. John Etchemendy of Stanford University, Dr. Carolyn Talcott of SRI, Dr. RV Guha of Microsoft, Dr. Alan Kay of Viewpoints Research, Harsh Sikka of Manifold Computing, and Dr. Navin Budhiraja and Dr. Sanjay Rajagopalan of Vianai Systems for valuable advice and criticism, Xinrui Wang of Vianai Systems for valuable contribution and measurements, and Sh. Manoj Modi of Reliance Industries for a stimulating conversation which inspired this work.